\documentclass[10pt,twocolumn,letterpaper]{article}

\usepackage{cvpr}
\usepackage{times}
\usepackage{epsfig}
\usepackage{multirow}
\usepackage{graphicx}
\usepackage{amsmath}
\usepackage{gensymb}
\usepackage{amssymb}
\usepackage{afterpage,float}
\usepackage{tabularx,ragged2e,booktabs,caption}
\newcolumntype{C}[1]{>{\Centering}m{#1}}

\usepackage{sidecap}
\graphicspath{{figures/}}

\usepackage[pagebackref=true,breaklinks=true,letterpaper=true,colorlinks,bookmarks=false]{hyperref}

\cvprfinalcopy 

\def\cvprPaperID{2714} 

\ifcvprfinal\fi
\begin{document}

\title{Deep representation learning for human motion prediction and classification}

\author{
  Judith B\"utepage$^1$ \quad  Michael J.~Black$^{2}$ \quad Danica Kragic$^1$ \quad Hedvig Kjellstr\"om$^{1}$ \\ \and
  {\centering{$^1${Department of Robotics, Perception, and Learning, CSC, KTH, Stockholm, Sweden}}}
  \and
  {\centering{
 $^2${Perceiving Systems Department, Max Planck Institute for Intelligent Systems, T\"ubingen, Germany}}}
 \and
  {\centering{\tt\small  butepage@kth.se, black@tuebingen.mpg.de,  dani@kth.se, hedvig@kth.se } }
}

\maketitle

\begin{abstract}
   Generative models of 3D human motion are often restricted to a small number of activities and can therefore not generalize well to novel movements or applications.
   In this work we propose a deep learning framework for  human motion capture data that learns a generic representation from a large corpus of motion capture data and generalizes well to new, unseen, motions. Using an encoding-decoding network that learns to predict future 3D poses from the most recent past, we extract a feature representation of human motion. Most work on deep learning for sequence prediction focuses on video and speech. Since skeletal data has a different structure, we present and evaluate different network architectures that make different assumptions about time dependencies and limb correlations. To quantify the learned features, we use the output of different layers for action classification and visualize the receptive fields of the network units. Our method outperforms the recent state of the art in skeletal motion prediction even though these use action specific training data. Our results show that deep feedforward networks, trained from a generic mocap database, can successfully be used for feature extraction from human motion data and that this representation can be used as a foundation for classification and prediction.  
\end{abstract}

\thispagestyle{empty}

\section{Introduction}
\label{sec:intro}

An expressive representation of human motion is needed not only for action classification  but also motion prediction and generation. A general representation of the skeletal pose and motion can serve different purposes in different fields. In computer vision, an adequate representation of movements can facilitate tracking and recognition. In robotics, this representation can be used to map  human motion to the robot's embodiment. The representation can also build the foundation for inference of intention and interpretation of goal-directed actions. 
Thus, there is a need for a sufficient and efficient representation that is generalizable to novel movements and  with a high transferability factor to different applications. 
Furthermore, this representation needs to encode both the correlations between joints and limbs and the temporal structure of human motion. The aim of this work is to develop and investigate learned representations of skeletal human motion data that can be used in a variety of tasks and are not tuned towards specific motion patterns.
\begin{figure}[t!]
\begin{center}
   \includegraphics[width=0.9\linewidth]{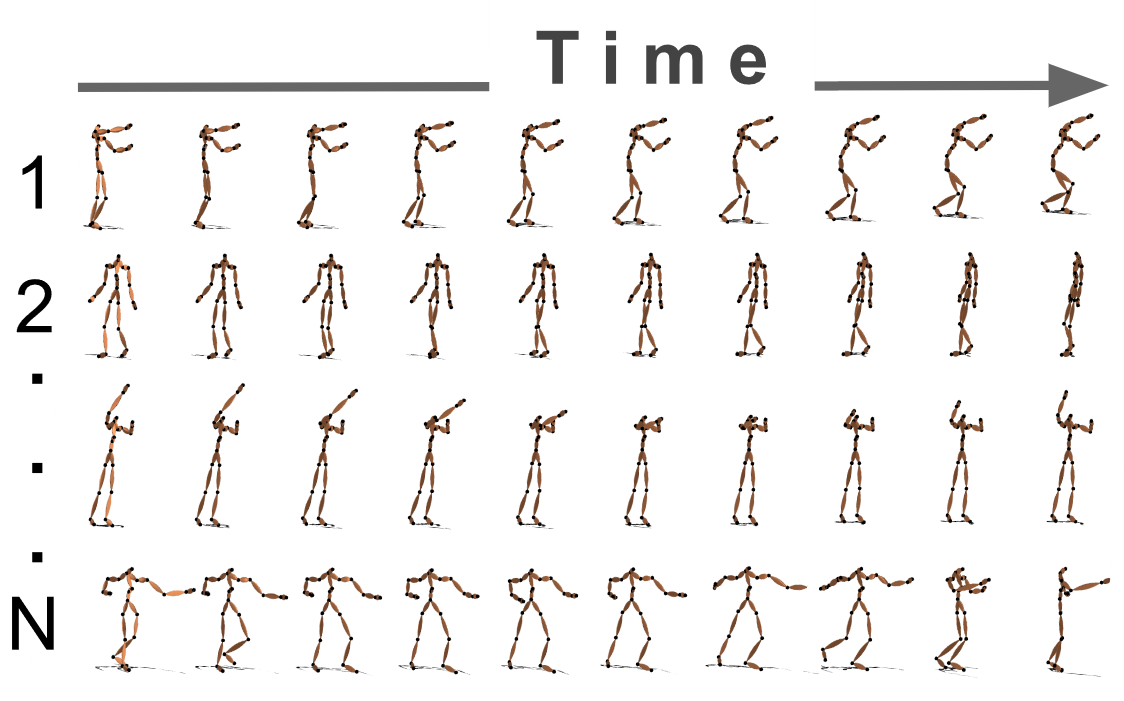}
\end{center}
\caption{The spike-triggered average pose \cite{scontimoncelli2004characterization} for different units in the middle layer of the hierarchical temporal encoder. Each sequence covers a duration of 1600 ms.  
} \label{fig:front} 
\end{figure}

Deep neural networks (DNN) have been found to automatically learn features that can generalize to novel tasks depending on the structure of the network and the tasks at hand \cite{azizpour2015generic}.
The advantage of deep architectures over shallow functions has been  attributed to the ability to uncover sparse, distributed representations in a hierarchical manner \cite{bengio2013representation}. Assuming that high-dimensional observed data points have been generated from a low-dimensional manifold, regularized autoencoders can be used to approximate the data-generating density locally \cite{alain2014regularized}. Additionally, convolutional neural networks (CNN) combine local feature extraction, weight sharing and pooling to extract invariant, increasingly complex features \cite{lecun1998gradient}. 

In order to capture correlations in temporal data such as video recordings, 3D CNNs have been proposed which apply convolutions in both temporal and spatial directions \cite{ji20133d}. However, state-of-the art convolutional techniques are not directly applicable to human motion capture (mocap) data. The local structure of images results in meaningful filter responses. Due to the hierarchical structure of the human body this does not directly apply to mocap data. The joints within a limb correlate over time while the joints of different limbs might be highly uncorrelated. To capture this correlation, the convolutional filters need to cover the whole range of joints such that convolution only occurs in temporal direction.  

A different approach to sequence learning are recurrent networks, such as Long Short-Term Memory networks (LSTM) \cite{hochreiter1997long} in which network units have recurrent connections such that information about previous activations can be propagated over time. While being well-suited for periodic data, recurrent networks perform less well when confronted with aperiodic time series. Although some human motion patterns, such as walking, are highly periodic, many more complex movements do not fall into this category. Due to this, most recent approaches train a separate model for each human action. This seriously limits their generalization to novel motions, actions, and tasks. Furthermore, these models are known to be of higher computational complexity than feed-forward networks and are difficult to train \cite{pascanu2013difficulty}.

In this work, we propose fully-connected networks with a bottleneck that learn to predict a number of future mocap frames given a window of previous frames. Thus, we train a temporal encoder of human motion. Due to the structure of the data, we hypothesize that fully-connected encoders are more expressive than state-of-the-art CNN architectures. Instead of a recurrent structure, we directly pass the recent history to the model, thus avoiding the difficulties of training recurrent networks and their tendency towards periodic motion.
We investigate how two different structural priors affect the representation. The first prior encodes different time scales by convolution over time. The second prior encodes the hierarchical structure of the human body with help of a fully-connected graph network.  In the experiments we firstly visualize the learned feature representation and secondly compare our models to state-of-the art models for human motion  action classification and prediction.

The main contributions of this work are:  
\begin{enumerate}
\itemsep-0.5em 
\item We develop an unsupervised representation learning scheme for long-term prediction of everyday human motion that is not confined to a small set of actions.
\item We train the model on a large portion of the CMU mocap dataset, producing a generic representation.
\item We demonstrate that our learned low-dimensional representation can be used for action classification and that we outperform more complex deep learning models in terms of motion prediction.
\item Our approach can be viewed as a generative model that has low computational complexity once trained, which makes it suitable for online tasks.
\end{enumerate}

\section{Related work}
\label{sec:related}

We focus our review of related work to that concerned with skeletal action recognition and human motion prediction and synthesis. Historically, many approaches have been based on hand-crafted features or joint correlation patterns \cite{han2016space}. Here we focus mainly on recent deep learning approaches most related to our method. 

In order to guarantee accurate action recognition, not only the human pose but also the trajectory over time need to be taken into account. In \cite{wu2014leveraging}, the prominent use of Hidden Markov Models (HMMs) is combined with multilayer perceptrons to model action-dependent hidden state trajectories. For this, the observed Cartesian skeleton data is taken as the input to the network, which predicts a hidden state feature vector that is trained to represent the current action in a supervised fashion. Thus, the evolution of actions over time can be classified. In contrast to this work, we do not force the latent representation to align with actions but rely solely on unsupervised learning to keep the representation as general as possible. 

In a different approach a hierarchical RNN is employed to directly classify actions from Cartesian skeleton data in a supervised manner \cite{du2015hierarchical}. The layers of this hierarchy are bidirectional RNNs, which successively receive information from more limbs the higher they are positioned in the network. The focus lies solely on action recognition with help of temporal dynamics. Instead, we aim at representation learning and prediction and use recognition mainly as a validation tool.  

Most comparable to our approach is the work described in \cite{liu2014feature} which proposes a deep sparse autoencoder (DSAE) for mocap data. The model is trained to reconstruct 0.2 seconds of subsequent mocap frames on three recordings containing seven distinct motion sequences. For validation, the same seven movements are classified with random forests and support vector machines based on the output features of the middle layer. In contrast, we aim at a representation of general motion, considering a large variety of everyday actions. 
Moreover, our method is shown in Section \ref{sec:experiments_class} to outperform theirs.

In addition
to action classification, several groups have addressed the problem of motion synthesis and prediction. In an early work, Taylor et al.~\cite{taylor2006modeling} present an autoregressive Restricted Boltzmann Machine with binary hidden variables for human motion prediction. The experiments are restricted to walking, jogging and running motions. Instead, we seek a more general model that can capture a large variety of actions.

In \cite{holden2015deep}, a low-dimensional manifold of human motion is learned using a one-layer convolutional autoencoder. For motion synthesis, the learned features and high-level action commands form the input to a feed-forward network that is trained to reconstruct the desired motion pattern. While the idea of manifold learning resembles our approach, the use of convolutional and pooling layers prevents the implementation of deeper hierarchies due to blurring effects  \cite{holden2015deep}.  

An encoding scheme is also applied by \cite{fragkiadaki2015recurrent}, who use an encoder-recurrent-decoder (ERD) model to predict human motion amongst others. The encoder-decoder framework learns to reconstruct joint angles, while the recurrent middle layer represents the temporal dynamics. As the whole framework is jointly trained, the learned representation is tuned towards the dynamics of the recurrent network and might not be generalizable to new tasks.

Finally, a combination of recurrent networks and the structural hierarchy of the human body for motion prediction has been introduced by \cite{jain2015structural} in form of structural RNNs (S-RNN). By constructing a structural graph in which both nodes and edges consist of LSTMs, the  temporal dynamics of both individual limbs and the whole body are modelled. Without the aid of a low-dimensional representation, a single model is trained for each motion. Thus, the computational and model complexity of this approach are comparably high.

In contrast to previous work, we develop a simple representation learning scheme of human motion dynamics, 
which is shown (Section \ref{sec:mpe}) to outperform the state-of-the-art methods in motion prediction, and also enable prediction of a wider range of motions (Section \ref{sec:mpg}) than earlier work.
 As we imagine the extracted features to be generalizable to application tasks such as motion prediction in human-robot interaction, we require a robust and fast system that circumvents the pitfalls of convolutional and recurrent networks. The details of this approach are described below. 
\section{Methodology}

In this section we introduce our temporal encoding scheme in mathematical terms. Furthermore, we describe three variations of this model: symmetrical encoding, time-scale encoding and structural encoding. 

\subsection{Data processing and representation}
\label{sec:processing}
As in \cite{holden2015deep}, we represent the mocap skeleton in the Cartesian space, i.e., a frame at time $t$ is given by $\mathbf{f}_t = [f_{x,i,t},f_{y,i,t},f_{z,i,t}]_{i = 1:N_{joints}}$, of dimension  $3 \times  N_{joints} $ where $N_{joints}$ is the number of joints.  

For normalization purposes, we convert the joint angles into the Cartesian coordinates of a standardized body model \cite{SMPL:2015}. The joint positions are centred around the origin of the coordinate system, i.e. we disregard translation while the global rotation of the skeleton is preserved. For each recorded subject and trial we subtract the mean pose over the whole trial.  

A single time window of size $\Delta t$ is given by the respective number of data frames which are concatenated into a matrix $\mathbf{F}_{t:(t + \Delta t -1)} = [\mathbf{f}_t,\mathbf{f}_{t+1},\ldots,\mathbf{f}_{t+\Delta t-1}] $ of dimension  $3 \times N_{joints}  \times \Delta t$. The dataset consists of an input frame window $\mathbf{F}_{(t-\Delta t+1):t}$ and an output frame window $\mathbf{F}_{(t + 1):(t + \Delta t)}$ for each time step $t \in [\Delta t,(T-\Delta t-1)]$, where $T$ is the length of the recording.

\subsection{Temporal encoder}

Encoding-decoding frameworks commonly aim at uncovering a projection of high-dimensional input data onto a low-dimensional manifold and to subsequently predict output data based on this projection. Autoencoders constitute a well-known subcategory of these frameworks. Given the high-dimensional input data $\mathbf{x} \in \mathbb{R}^N$, autoencoders optimize 
\begin{equation}
\min_{f,g}|| \mathbf{x} - f(g(\mathbf{x}))||,
\end{equation}
where the encoder $\mathbf{y} = g(\mathbf{x})$ maps the input data into a low-dimensional space $\mathbf{y} \in \mathbb{R}^M, N>M$, and the decoder $\hat{\mathbf{x}} = f(\mathbf{y})$ maps back into the input space $\mathbf{\hat{x}} \in \mathbb{R}^N$. In general, the functions $f$ and $g$ are represented by symmetric multilayer perceptrons.

In this work, we propose an alternative approach in order to capture the temporal correlations of human motion data rather than a static representation of human poses. In a general manner, let $\mathbf{x}_{t} \in \mathbb{R}^{N}$  be an observation at time $t$ and $\mathbf{X}_{(t-\Delta t+1):t}  = [\mathbf{x}_{t-\Delta t+1}, \mathbf{x}_{t-\Delta t+2},\ldots,\mathbf{x}_{t}] \in \mathbb{R}^{N \times \Delta t}$ be a matrix that consists of the last $\Delta t$ observations at time $t$. Similarly, let 
$\mathbf{X}_{(t+1):(t+\Delta t)} = [\mathbf{x}_{t+1}, \mathbf{x}_{t+2},\ldots,\mathbf{x}_{t+\Delta t}] \in \mathbb{R}^{N \times \Delta t}$ be the matrix that contains the future $\Delta t$ observations at time $t$. Then a temporal encoder (TE) optimizes
\begin{equation}
\min_{f,g}||\mathbf{X}_{(t+1):(t+\Delta t)} - f(g(\mathbf{X}_{(t-\Delta t+1):t}))||,
\end{equation}
where the encoder $\mathbf{y} = g(\mathbf{X}_{(t-\Delta t+1):t})$ maps the input data into a low-dimensional space $\mathbf{y} \in \mathbb{R}^M, (N \times \Delta t)>M$, and the decoder $\hat{\mathbf{X}}_{(t+1):(t+\Delta t)} = f(\mathbf{y}) \in \mathbb{R}^{N \times \Delta t}$ maps back into the data space. Instead of a purely symmetric setting, the functions $f$ and $g$ can be differently structured. While the encoder has to take local features into account, the decoder needs to learn a globally valid structure. 

In our application, the input and output matrices are of dimension $3 \times N_{joints}  \times \Delta t$ such that the encoder $\mathbf{y} = g(\mathbf{F}_{(t-\Delta t+1):t})$ maps the input data into a low-dimensional space $\mathbf{y} \in \mathbb{R}^M, (3 \times  N_{joints}  \times \Delta t)>M$, and the decoder $\hat{\mathbf{F}}_{(t+1):(t+\Delta t)} = f(\mathbf{y}) \in \mathbb{R}^{3 \times  N_{joints}   \times \Delta t}$ maps back into the data space.

\begin{figure}[t]
\begin{center}
   \includegraphics[width=0.8\linewidth]{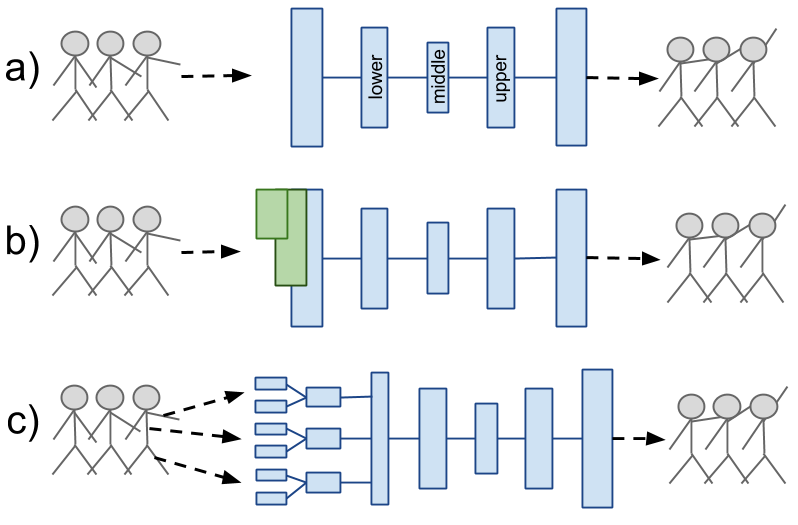}
\end{center}
   \caption{The structure of three different temporal encoders that encode the last time frames (left) and reconstruct the next time frames (right) of a skeletal movement, here lifting an arm. The number and size of layers is only for illustrative purposes. Blue layers represent fully-connected layers while green layers represent convolutional layers that convolve only in the direction of time. a) {\bf S-TE}: A symmetric structure for encoder and decoder. b) {\bf C-TE}: The encoder considers different time scales. c)  {\bf H-TE}: The hierarchy of the human body is directly incorporated by the encoder.}
\label{fig:networks}
\end{figure}

\subsection{Network structure}

As depicted in Figure \ref{fig:networks}, in this work we present three different temporal encoder structures: symmetric coding, time-scale encoding and hierarchy encoding.

\noindent \textbf{Symmetric coding.}
The symmetric structure as shown in Figure \ref{fig:networks} a) follows the general idea of autoencoders. As the decoder is a mirrored version of the encoder, the decoder can be viewed as an approximation of the inverse of the encoder. In later sections, this approach will be denoted by \textit{symmetric temporal encoder} (S-TE).

\noindent \textbf{Time-scale encoding.}
As human motion can be described on different time scales, this property can be explicitly introduced to the temporal encoder. While convolution over joints is impractical as discussed in Section \ref{sec:intro}, filters that cover the whole range of joints can be convolved in time direction. Thus, for a given window size $\Delta t^w$, the convolutional filters are of size $3 \times N_{joints} \times \Delta t^w$, where $3$ indicates the three dimensions in Cartesian space $x,y \text{ and } z$. The input data is convolved with filters of different sizes. The output of these convolutional layers is concatenated and further processed by fully-connected layers in a encoder-decoder fashion, as illustrated in Figure \ref{fig:networks} b). In later sections, this approach will be denoted by \textit{convolutional temporal encoder} (C-TE).

\noindent \textbf{Hierarchy encoding.}
The human body can be represented by a tree in which the nodes consist of the individual joints, connected to the nodes of corresponding limbs in the body.
Let this tree be composed of $L$ layers, where each layer $l \in [0,L-1]$ consists of $N_l$ nodes, denoted by $\nu_{l,i}, i \in [0,N_l-1]$. Each parent layer $l \in [1,L-1]$ is connected to its child layer $k = l - 1$ by a set of links. For node $i$ in layer $l$ and node $j$ in layer $k$ a link is denoted by $\xi_{(l,i), (k,j)}$.
In this work, we model these nodes as single feedforward layers in the temporal decoder that are selectively connected to their parent layers. 
Each node in the bottom layer receives input from a single joint, i.e. $N_0 = N_{joints}$.
Subsequently, these nodes are connected by a parent that represents a limb, i.e. $\xi_{(0,i), (1,j)} = 1$ if joint $i$ belongs to limb $j$, $\xi_{(0,i), (1,j)} = 0$ otherwise.  
In this manner the hierarchy is formed until a single node represents the entire body, see Figure \ref{fig:networks} c). This single layer serves as the input to the temporal encoder, which is trained jointly with the tree graph. In later sections, this approach will be denoted by \textit{hierarchical temporal encoder} (H-TE).

\section{Experiments}

\label{sec:experiments}

Our models are trained on 1035 recordings that are part of the CMU mocap database \cite{cmu}. This database contains 2235 recordings of 144 different subjects performing a large variety of complex movements. As a number of recordings have a sampling rate of 120 Hz, while others are sampled at 60 Hz, we sample the former trials down to 60 Hz. For the evaluation, we use recordings from the H3.6M dataset \cite{h36m_pami}, which are preprocessed as described above. The current models are trained with a time window of  $\Delta t = 100$ frames, or around 1660 ms. This enables substantially longer predictions compared to \cite{fragkiadaki2015recurrent} and \cite{jain2015structural}. 
In summary, the input and output data points consist of   $3 \times N_{joints}  \times \Delta t = 3 \times 24 \times 100 = 7200$ dimensions.

All models are implemented and trained using the Caffe deep learning framework  \cite{jia2014caffe}. In order to prevent overfitting and to keep the learned representation close to the human motion manifold, we apply an increasing amount of dropout noise to the data layer during training. Additionally, we apply layerwise pre-training which seems to decrease training time but not to have a significant affect on the final performance. Additional information about the structure of the networks and training details can be found in the supplementary material. 

\begin{figure}[t]
\begin{center}
   \includegraphics[width=0.8\linewidth]{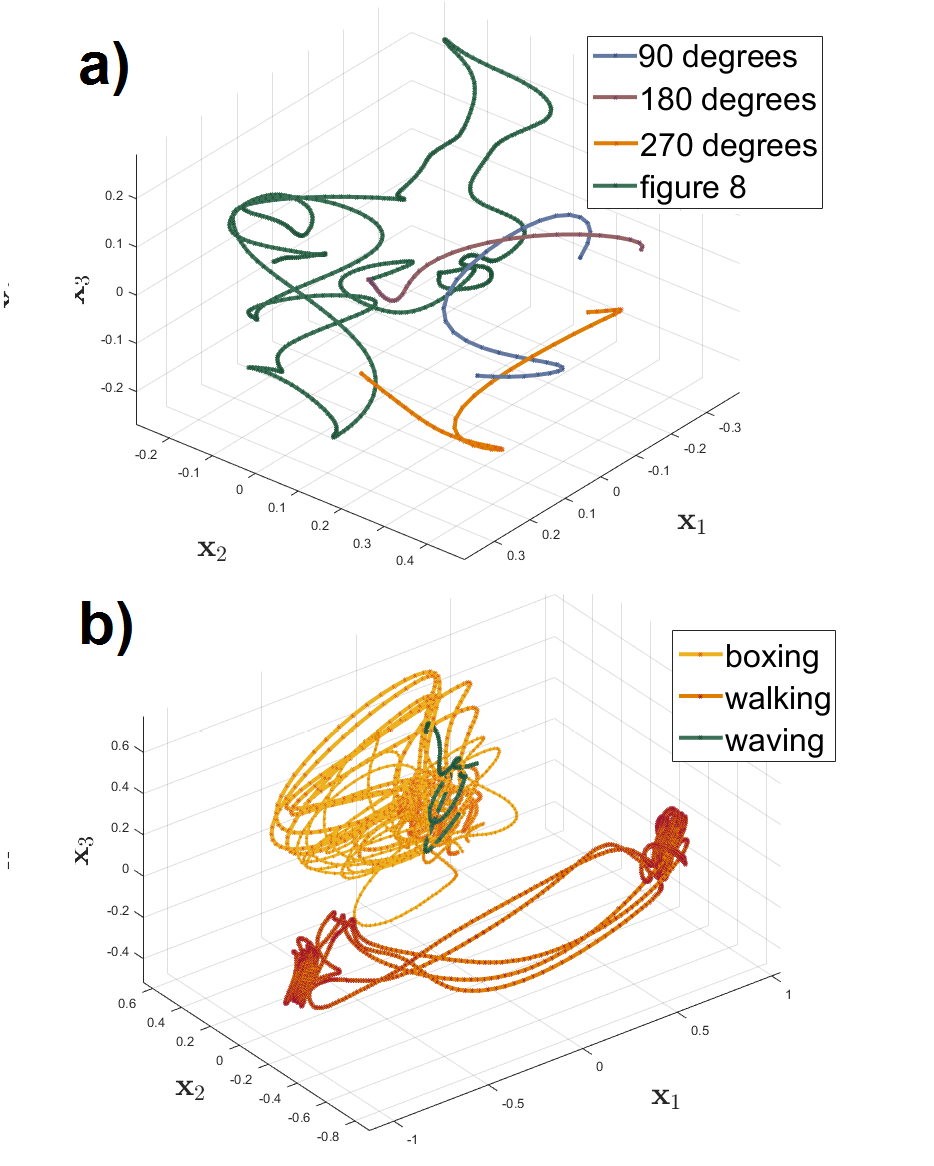}
\end{center}
   \caption{Three dimensional GPFA \cite{byron2009gaussian} of the feature dynamics in the middle layer of H-TE. a) Whole-body rotations of different degrees and the walking trajectory that resembles the "figure 8". b) Entire action sequences for boxing, waving and walking.  }
\label{fig:dyn}
\end{figure}


\subsection{Feature visualization}

Visualization of neural features has been mainly addressed for CNNs, see e.g. \cite{zeiler2014visualizing}. Visualization of features from modalities other than images has been less prominent. In this work, we apply methods from the area of computational neuroscience to examine the learned representation.

In Figure \ref{fig:front}, we present the average pose that excited a number of units in the middle layer of H-TE. This ``spike-triggered average'' \cite{scontimoncelli2004characterization} is computed by weighting input data points by the activity of a specific unit. In order to reduce the noise, we only consider poses and network activity when the output of the sigmoid unit exceeds 0.8. It becomes apparent that the units encode different motions. Both whole-body rotation and posture as well as single limb movements are represented. 

The goal of this work is to learn a low-dimensional representation of human motion dynamics that encodes the underlying action. Thus, data points that are similar in pose space should be close to each other in this low-dimensional space  and longer motion sequences should constitute a trajectory on this manifold. In order to verify whether this holds true for the  encoding of temporal dynamics learned by our models, we make use of Gaussian Process Factor Analysis (GPFA) \cite{byron2009gaussian}. GPFA is a dimensionality reduction technique that takes temporal structure into account. In comparison to Principal Component Analysis, GPFA can uncover non-linear correlations in temporal data. Originally used for the analysis of spike trains, we apply GPFA to the output of middle layers in our networks over motion segments of different actions from the CMU dataset. 

In Figure \ref{fig:dyn},  we depict different actions in the three main factor dimensions over time uncovered by GPFA. Whole-body turns to different degrees are shown in Figure \ref{fig:dyn} a). The dynamics expressed by the units seem to encode the degree of turning. While the length increases with more degrees of turning, the dynamics approach closed circles. The representation of the ``figure 8'' movement does resemble the two-loops structure of the motion trajectory. 

Figure \ref{fig:dyn} b) displays different actions. The circular pattern of \textit{walking} is clearly distinguishable from \textit{boxing} and \textit{waving}.  Additionally, the repetitive motion sequences in \textit{boxing} and \textit{waving} are reflected in the latent space. As both of these actions mostly concern arm movements, they are well separated from the \textit{walking} trajectory.

\begin{figure}[b!]
\captionof{table}{Action classification rate, CMU mocap dataset} \label{tab:mce}
\centering{ 
\resizebox{0.9\linewidth}{!}{
  \begin{tabular}{|l|c|c|c|}
    \hline
    \multicolumn{1}{|l|}{Method} & \multicolumn{3}{|c|}{Classification rate}    \\ \hline  
    \multicolumn{1}{|c|}{Data (1.6 s)}  & \multicolumn{3}{|c|}{0.76}    \\ \hline
    \multicolumn{1}{|c|}{PCA}  & \multicolumn{3}{|c|}{0.73}    \\ \hline
         & Lower Layer & Middle Layer & Upper Layer \\
    \hline  
     \multicolumn{1}{|c|}{DSAE \cite{liu2014feature}} & 0.72 & 0.65 & 0.62  \\ \hline
    S-TE & \textbf{0.78} &  0.74 & 0.67 \\
   \hline 
    C-TE & \textbf{0.78} &  0.74 & 0.73 \\
   \hline 
    H-TE & 0.77 &  0.73 & 0.69 \\
   \hline 
  \end{tabular}
  }
  }
\end{figure}

\subsection{Action classification}
\label{sec:experiments_class}

In order to evaluate the expressiveness of the learned features, we classify the underlying actions based on the output features of different layers. For each action, we extract the output features for every time step of the recording and store this together with the label of the action. Our classifier is a two-layer fully-connected neural network with a softmax output layer that is trained to classify the output features of a given layer. We present results for layers near to the data (lower layer), for the bottleneck layer (middle layer) and a layer near to the output (upper layer). For comparison, we additionally classify directly on the data points and the first 75 principal components extracted with PCA which explain 95 \% of the variance in the data. 

Classification rates for the CMU dataset have been reported by a number of groups, e.g., \cite{barnachon2014ongoing} and \cite{kadu2014automatic}. These methods concentrate on pure classification and report up to 99.6 \% accuracy \cite{kadu2014automatic}, while not being suitable for representation learning and generation of future motion. Therefore, we will here compare our results with a feature extraction approach,  deep sparse autoencoders, as described in \cite{liu2014feature}, see Section \ref{sec:related}. In order to make the results comparable, we train a DSAE on our dataset and adjust the number of parameters to be identical to the parameters of H-TE.   

\begin{figure*}[t!]
\begin{center}
   \includegraphics[width=0.85\linewidth]{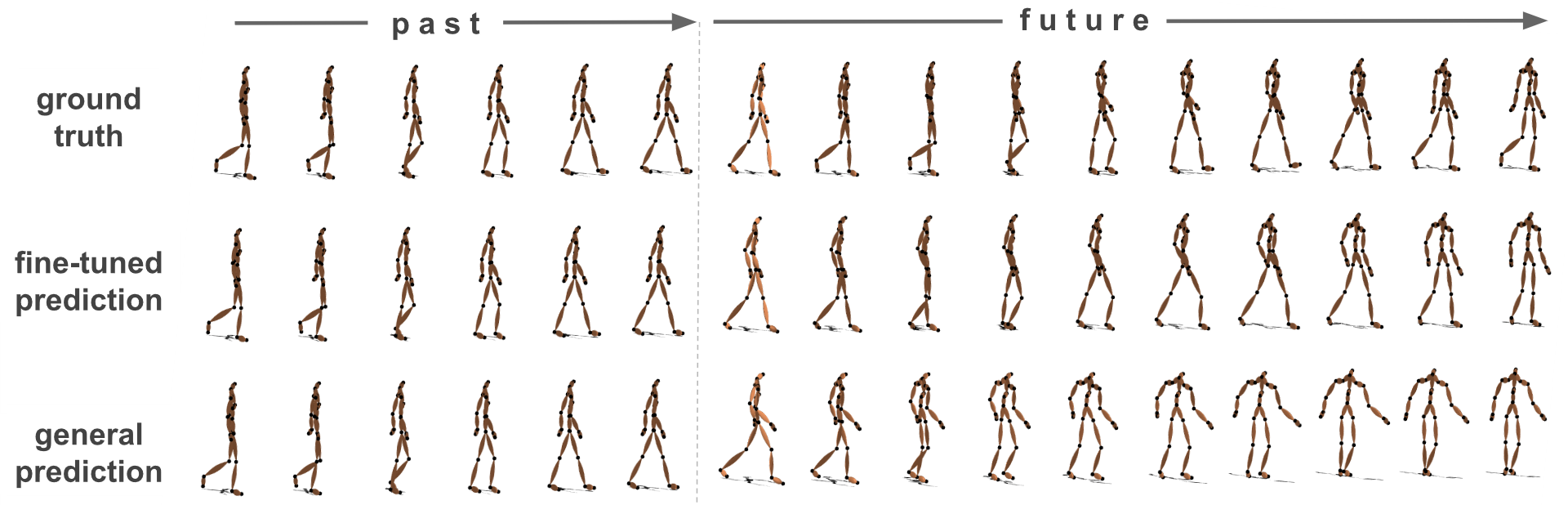}
\end{center}
   \caption{Prediction of a walking sequence from the H3.6M dataset. One second of the past movement is depicted together with 1600 ms predictions of future movement made by H-TE-F (middle) and H-TE (bottom).  }
\label{fig:walk}
\vspace{-1mm}
\end{figure*}

We classify whole action sequences instead of single movement sequences as in \cite{liu2014feature}. For this, we follow the experimental setup for CMU mocap action classification described in \cite{barnachon2014ongoing}. However, we perform cross-validation by training on the majority of the listed recordings and testing on the first 8 seconds of the remaining recordings to have a comparable measure for all actions. Thus, our reported results are the average classification rates for the actions \textit{walk, run, punching, boxing, jump, shake hands, laugh, drink} and \textit{eat}.  The results are shown in Table ~\ref{tab:mce}. 

It becomes apparent that a data sequence alone results in a better classification rate than the representation extracted with the DSAE. In comparison, our models show comparable or slightly higher classification rates.
This implies that the temporal encoding in comparison to a basic autoencoder extracts more relevant information from the data. While our data layer is of dimensionality 7200, the dimensionality of the lower and upper layers is 300 and the dimensionality of the middle layers is 100. Thus, the features of the low-dimensional layers reflect essential information about the performed actions contained in the data.

\subsection{Motion prediction of specific actions}
\label{sec:mpe}

Here, we compare the predictive power of our three models S-TE, C-TE and H-TE to the recently proposed ERD \cite{fragkiadaki2015recurrent} and S-RNN \cite{jain2015structural} models and, following their example, a 3-layered LSTM (LSTM3L). We evaluate our models on the H3.6M dataset \cite{h36m_pami}.  For this, we use pre-trained versions of the recurrent models and implementations made publicly available by  \cite{jain2015structural}.  All of these models have been trained with recordings from the H3.6M dataset. These were down sampled to 25 Hz and joint angles were converted into exponential maps. As our time window covers approximately 1660 ms, the recurrent networks are initialized with 40 frames, which corresponds to 1600 ms. For each action, a separate pre-trained, recurrent model is used. In order to make the two approaches comparable, we convert the exponential map predictions into the Cartesian space as described in Section \ref{sec:processing}. However, global rotation and translation are set to zero as the models have been trained without this information.  

Note that in contrast to the recurrent networks our models were not trained on the H3.6M  dataset. In order to test action specific performance, we fine-tune H-TE to each of the tested actions and report the results separately, denoted by H-TE-F. For this, the training subjects are S1, S6, S7 and S8 and the test subject is S5.

\begin{figure}[bh!]
\captionof{table}{Motion prediction error, single actions} \label{tab:mpe} 
\centering{ 
\resizebox{0.9\linewidth}{!}{
  \begin{tabular}{|l|c|c|c|c|c|}
    \hline
    \multirow{2}{*}{Method} &
      \multicolumn{3}{c}{Short Term} &
      \multicolumn{2}{|c|}{Long Term} \\   
     & 80ms & 160ms & 320ms & 560ms & 1000ms \\
     \hline  
    \multicolumn{6}{|c|}{\textbf{Walking}}  \\  
    \hline  
    ERD \cite{fragkiadaki2015recurrent} & 0.18 & 0.23 & 0.34 & 0.45 & 0.57  \\
    \hline
    S-RNN \cite{jain2015structural} &  0.18 & 0.21 & 0.29 & 0.41 & 0.53   \\
    \hline
    LSTM3L & 0.18 & 0.23 & 0.32 & 0.39 & 0.43  \\
    \hline
    S-TE & 0.33 & 0.35 & 0.37 & 0.37 & 0.4 \\
    \hline
    C-TE & 0.18 & 0.2 & 0.26 & 0.32 & 0.36 \\
    \hline
    H-TE & \textbf{0.17} &   \textbf{0.18}&  \textbf{0.23}&  \textbf{0.28}&  \textbf{0.31}  \\
   \hline  
   H-TE-F & \textbf{0.16} &   \textbf{0.17}&  \textbf{0.2}&  \textbf{0.24}&  \textbf{0.24}  \\
   \hline 
    \multicolumn{6}{|c|}{\textbf{Smoking}}  \\  
    \hline 
    ERD \cite{fragkiadaki2015recurrent} & 0.35 & 0.39 & 0.43 & 0.49 & 0.58   \\
    \hline
        S-RNN \cite{jain2015structural} & 0.33 & 0.36 & 0.42 & 0.5 & 0.57   \\ 
    \hline
    LSTM3L & \textbf{0.26} & 0.3 & 0.37 & 0.42 & 0.48 \\
    \hline
     S-TE & 0.4 & 0.4 & 0.4 & 0.42 & 0.49   \\
    \hline
    C-TE & \textbf{0.26} & 0.27 & 0.33 & 0.4 & 0.49 \\
    \hline
    H-TE & \textbf{0.26}  &  \textbf{0.26} &  \textbf{0.29}&  \textbf{0.35}&  \textbf{0.41}  \\
    \hline  
     H-TE-F & \textbf{0.17}  &  \textbf{0.17} &  \textbf{0.19}&  \textbf{0.23}&  \textbf{0.27}  \\
    \hline  
    \multicolumn{6}{|c|}{\textbf{Eating}}  \\  
    \hline 
    ERD \cite{fragkiadaki2015recurrent} & 0.23 & 0.27 & 0.34 & 0.42 & 0.52    \\
    \hline
    S-RNN \cite{jain2015structural}& 0.18 & 0.23 & 0.32 & 0.41 & 0.41  \\
    \hline
    LSTM3L &  \textbf{0.17} & 0.23 & 0.32 & 0.37 & 0.41  \\
    \hline
     S-TE & 0.33 & 0.34 & 0.35 & 0.37 & 0.42 \\
    \hline
    C-TE & 0.19& 0.21 &  0.25& 0.31 &  0.37 \\
    \hline
    H-TE & 0.2 & \textbf{0.2}& \textbf{0.23} & \textbf{0.29} &  \textbf{0.37}  \\
    \hline
     H-TE-F& \textbf{0.15} & \textbf{0.15} & \textbf{0.17} & \textbf{0.21} &  \textbf{0.26}  \\
    \hline
    \multicolumn{6}{|c|}{\textbf{Discussion}}  \\  
    \hline 
    ERD \cite{fragkiadaki2015recurrent} & 0.29 & 0.34 & 0.42 & 0.46 & 0.5    \\
    \hline
    S-RNN \cite{jain2015structural} & 0.35 & 0.37 & 0.48 & 0.55 & 0.54 \\
    \hline
    LSTM3L & 0.44 & 0.46 & 0.54 & 0.56 & 0.57   \\
    \hline
     S-TE & 0.22 & 0.23 & 0.32 & 0.26 & 0.27 \\
    \hline
    C-TE & 0.15& 0.17 &  0.2& 0.25 &  0.31\\
    \hline
    H-TE & \textbf{0.16} & \textbf{0.17} & \textbf{0.2} & \textbf{0.22} &  \textbf{0.24}  \\  
    \hline
    H-TE-F & \textbf{0.13} & \textbf{0.14} & \textbf{0.18} & \textbf{0.2} &  \textbf{0.22}  \\  
    \hline
  \end{tabular}
  }
  }
\end{figure}

\begin{figure}[t!]
\begin{center}
   \includegraphics[width=0.85\linewidth]{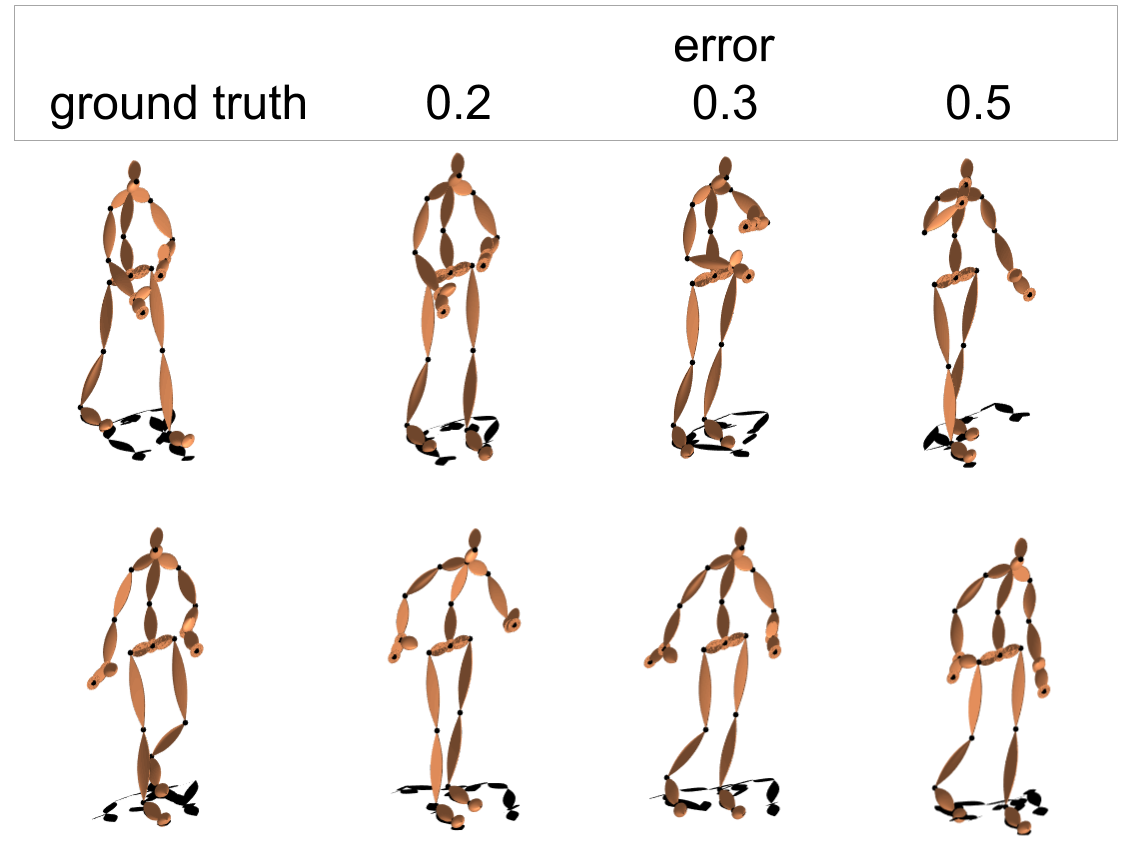}
\end{center}
   \caption{A visual illustration of error rates. Two different poses originating from the actions \textit{eating} (top row) and \textit{walking} (bottom row) are depicted together with poses that result in the error rates 0.2, 0.3 and 0.5 w.r.t. the ground truth. }
\label{fig:error}
\vspace{-2.5mm}
\end{figure}

Following \cite{jain2015structural} we evaluate the predictive power of the models on the actions \textit{walking, smoking, eating} and \textit{discussion} for short-term predictions of 80 ms, 160 ms and 320 ms and long-term predictions of 560 ms and 1000 ms. For this, we compute the Euclidean distance between the ground truth and the prediction made by each model for a given frame and normalize with the number of joints, resembling the mean squared error over joints. The values reported here are the average error over eight randomly selected sequences of each action. The results are presented in Table \ref{tab:mpe}. Figure \ref{fig:error} illustrates different error rates visually. 

While LSTM3L outperforms some of our models for the initial predictions, the temporal encoders show better performance for predictions of 160 ms and more. Since the encoders are trained to jointly predict an entire time window, they suffer less from diffusion and propagated errors.

Because the action \textit{``discussion''} is a complex, non-stationary action, the recurrent networks struggle to make short-term predictions. In contrast, our models are able to infer future frames. Interestingly, the symmetric temporal encoder S-TE and the convolutional temporal encoder C-TE are outperformed by the hierarchical temporal encoder H-TE in most predictions. This indicates that a structural prior is beneficial to motion prediction. As expected, the fine-tuning to specific actions decreases the prediction error and is especially effective during long-term prediction and for actions that are not contained in the original training data, such as   \textit{``smoking''}.

We depict the prediction for a walking sequence contained in the H3.6M dataset for the whole range of around 1600 ms in Figure \ref{fig:walk}. The fine-tuned model (middle) predicts the ground truth (top) with a high level of accuracy. The prediction by the general model is accurate up to around 600 ms. Note that predictions over 560 ms can diverge from the ground truth substantially due to stochasticity  in human motion \cite{fragkiadaki2015recurrent} while remaining meaningful to a human observer. 

\subsection{General motion prediction}
\label{sec:mpg}
In order to test how well our models generalize to unseen data, we present the average forecast error made for all recordings of subject S1, S5, S6, S7 and S8 contained in the H3.6M dataset. For this, we slide a window over each  recording and make a prediction for every time step. The forecast errors presented in Table \ref{tab:mpegm}  are averaged over these predictions for all recordings of all subjects.  Note that our models were trained on the CMU mocap database. Thus, the H3.6M dataset poses the challenge of novel subjects and actions. For comparison, we also present the average forecast error made on our held out testset (15 \% of the data) of the CMU database. As the recurrent networks are tuned towards specific actions, they are not able to generalize to the same extent as our models. In Table \ref{tab:mpegm} we present the average prediction error made for the four actions presented in Section \ref{sec:mpe} by all four action-specific models for each recurrent model. 

The general performance of out models stays close to the results presented for single actions in  Section \ref{sec:mpe}, while the recurrent models generalize less. Interestingly, the C-TE outperforms the H-TE for short-term predictions while its performance for long-term prediction approaches S-TE. As the convolutions of C-TE take different time scales into account, this approach concentrates on local temporal information. In contrast, the S-TE encodes global information about the entire input data and is therefore more likely to make accurate long-term predictions.

\begin{figure}[h!]
\captionof{table}{Motion prediction error, general motion} 
\centering{ 
\resizebox{1\linewidth}{!}{
  \begin{tabular}{|l|c|c|c|c|c|c|}
    \hline
    \multirow{2}{*}{Method} &
      \multicolumn{3}{c|}{Short Term} &
      \multicolumn{3}{c|}{Long Term} \\   
     & 80ms & 160ms & 320ms & 560ms & 1000ms &  1600ms \\
     \hline  
     \multicolumn{7}{|c|}{\textbf{H3.6M (average over four action-specific models) }}  \\  
    \hline  
    ERD [8] & 0.39  & 0.44  & 0.53  & 0.6  & 0.67  & 0.7 \\
    \hline
    S-RNN [13] &  0.34 & 0.38 & 0.45  & 0.52 & 0.59  & 0.63  \\
    \hline
    LSTM3L &  0.27 & 0.33 & 0.42  & 0.49 & 0.57  & 0.62  \\
     \hline  
    \multicolumn{7}{|c|}{\textbf{H3.6M}}  \\  
    \hline
    S-TE & 0.36 & 0.37 & 0.37 & 0.38 & 0.43  & 0.45 \\
    \hline
    C-TE & \textbf{0.21} & 0.23 & 0.27 & 0.34 & 0.42  & 0.45 \\
    \hline
    H-TE & \textbf{0.21} &   \textbf{0.22}&  \textbf{0.25}&  \textbf{0.3}&  \textbf{0.36} &\textbf{0.39}  \\
   \hline  
   \multicolumn{7}{|c|}{\textbf{CMU}}  \\  
    \hline  
 
    S-TE & 0.3  & 0.31 & 0.33 & 0.35 & 0.37 & 0.37 \\
    \hline
    C-TE & \textbf{0.18} & 0.21 & 0.25 & 0.30 & 0.33  & 0.35 \\
    \hline
    H-TE & \textbf{0.18} &   \textbf{0.20}&  \textbf{0.24}&  \textbf{0.28}&  \textbf{0.31} &\textbf{0.33}  \\
   \hline

  \end{tabular}
 }} \label{tab:mpegm} 
\end{figure}

\subsection{Missing data}

In realistic applications, the representation uncovered by the temporal encoders needs to be both general and robust towards noise and missing frames in the input data. Thus, the models should be able to infer the position of limbs with missing input data by relying on the learned correlations in the training data. We test this hypothesis by setting the data of all joints belonging to the same limb over the entire input window equal to zero. Upon visual inspection, as illustrated in Figure \ref{fig:missing}, it becomes apparent that the model is able to fill in the missing information. Especially in the case of a missing arm, in Figure \ref{fig:missing} b), the model is able to predict that both arms are raised in future time steps. 

\begin{figure}[t!]
\captionof{table}{Motion prediction error, missing data} \label{tab:mpemd} 
\centering{ 
\resizebox{0.9\linewidth}{!}{
  \begin{tabular}{|l|c|c|c|c|c|}
    \hline
    \multirow{2}{*}{Method} &
      \multicolumn{3}{c}{Short Term} &
      \multicolumn{2}{|c|}{Long Term} \\   
     & 80ms & 160ms & 320ms & 560ms & 1000ms \\
      \hline  
     \multicolumn{6}{|c|}{\textbf{Eating}}  \\  
     \hline
     S-TE & 0.33 & 0.34 & 0.35 & 0.37 & 0.42 \\
    \hline
    C-TE & \textbf{0.19}& 0.21 &  0.25& 0.31 &  0.37 \\
    \hline
    H-TE & 0.2 & \textbf{0.2} & \textbf{0.23} & \textbf{0.29} &  \textbf{0.37}  \\
    \hline  
    \multicolumn{6}{|c|}{\textbf{Eating (right arm missing)}}  \\  
    \hline   
     S-TE & 0.44&  0.44&  0.44&  0.46&  0.5 \\
    \hline
    C-TE & 0.31 &  0.33 & 0.36 &  0.4 &  0.46 \\
    \hline
    H-TE &  0.3 &  0.31 & 0.33 &  0.37 &  0.42 \\
    \hline  
    \multicolumn{6}{|c|}{\textbf{Eating (left leg missing)}}  \\  
    \hline   
     S-TE & 0.41&  0.41&  0.42&  0.42&  0.47 \\
    \hline
    C-TE & 0.3 & 0.31 &  0.35&  0.4& 0.5 \\
    \hline
    H-TE &  0.31 &  0.31 & 0.33 &  0.37 &  0.43 \\
    \hline   
  \end{tabular}
  }
  }
\end{figure}

\begin{figure}[b!]
\begin{center}
   \includegraphics[width=1\linewidth]{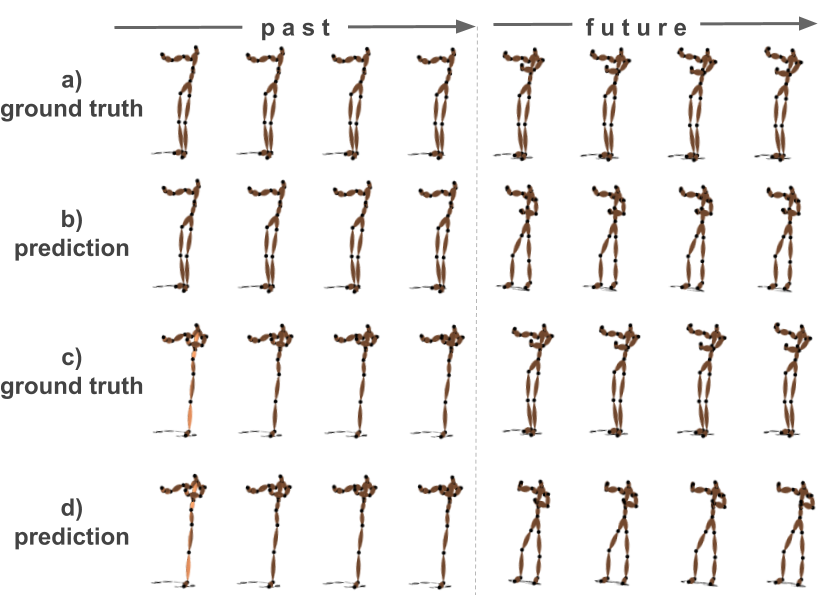}
\end{center}
   \caption{Prediction with missing data. Past and future time steps of the action \textit{taking photos}, a recording in the H3.6M dataset, with a missing arm and a missing leg in the input data. Both past and future consist of around 640 ms. a) Ground truth for a missing arm. b) Prediction by H-TE for a missing arm. c) Ground truth for a missing leg. d) Prediction by H-TE for a missing leg.} 
\label{fig:missing}
\end{figure}

To test these predictions quantitatively, we measured the average prediction error as described in Section \ref{sec:mpe}. In Table \ref{tab:mpemd}, we list the error of our models for a missing right arm and a missing left leg during "\textit{eating}". More results are presented in the supplementary material. In general, we observe that the error increases around one decimal for all models and all prediction times. However, compared to the errors of the recurrent approaches listed in Table \ref{tab:mpe}, the error stays comparably low. Thus, the models are able to infer the pose of the missing limb and do not diverge significantly from the original motion.

\section{Discussion}
\label{sec:discussion}

In this work, we presented a temporal encoder scheme for feature learning of human motion. Our main objective was to uncover a robust and general representation of human motion that can be used both as a generative model and as a feature extractor. We presented three approaches to this problem, all based on the idea of bottleneck encoding-decoding from past to future frames. The visualization of the learned representation shows that the layers encode a diverse range of motion in a structured, lower dimensional space. Due to this structure, action classification directly on the features without fine-tuning becomes possible. We demonstrated that our feed-forward networks outperform recurrent approaches for short-term and long-term predictions and that the predictions generalize to novel subjects and actions. Finally, the ability to infer the position of missing limbs indicates the robustness of our approach.

The performance of our feed-forward temporal encoders on these tasks can be ascribed to the simplicity of the approach and the bottleneck structure that forces the networks to learn an efficient and sufficient data representation.

While feed-forward networks require a pre-specified input window, one argument in favour of recurrent networks is that they are able to encode information over longer periods of time. However, they are more complex and appear to be less general and robust than pure feed-forward connections. As skeletal human pose data is low-dimensional compared to e.g. images, the training with long time windows does not pose a computational challenge. In this work, we made use of this fact and demonstrated that long-term predictions based on a sliding window are more accurate than recurrent approaches. 

The difference between the performances of our three models --  symmetric, convolutional and hierarchical -- might be influenced by the number of parameters and the structure of each network. In order to get a proper understanding of how these two factors interact, further investigations are needed. Additionally, a general observation for all models is that the prediction error increases for long-term predictions. In realistic applications, a measure of uncertainty for predictions might be required, such as provided by e.g. conditional variational autoencoders \cite{sohn2015learning}. In future work we plan to extend our approach to encode this information and to systematically investigate the impact of the window size $\Delta t$. Furthermore, we plan to test its applicability in real-time applications on 2D and 3D skeletal data.



\section*{Acknowledgement}
This work was partly supported by the EU through the project socSMCs (H2020-FETPROACT-2014) and Swedish Foundation for Strategic Research. 

{\small
\bibliographystyle{ieee}
\bibliography{egbib.bib}
}

\end{document}